\documentclass[11pt, letter]{article}
\usepackage[utf8]{inputenc}
\usepackage[final]{microtype}
\usepackage[margin=1in]{geometry}

\usepackage{natbib}
\usepackage{authblk}

\usepackage{caption}
\usepackage{subcaption}

\usepackage[hidelinks]{hyperref}
\usepackage{url}

\usepackage{graphicx}

\usepackage[inline]{enumitem}
\setenumerate[1]{leftmargin=1.5cm} 

\usepackage{alltt}

\usepackage{booktabs}
\usepackage{multirow}

\usepackage[inkscapeformat=png]{svg}
\usepackage{svg}

\title{On Sarcasm Detection with OpenAI GPT-based Models}

\author[1]{Montgomery Gole}
\author[1]{Williams-Paul Nwadiugwu}
\author[1]{Andriy Miranskyy}

\affil[1]{Department of Computer Science, Toronto Metropolitan University, Toronto, Canada}
\affil[]{{\{mgole, williams.nwa, avm\}@torontomu.ca}}

\date{}

\begin{document}

\maketitle

\begin{abstract}
Sarcasm is a form of irony that requires readers or listeners to interpret its intended meaning by considering context and social cues. Machine learning classification models have long had difficulty detecting sarcasm due to its social complexity and contradictory nature.

This paper explores the applications of the Generative Pretrained Transformer (GPT) models, including GPT-3, InstructGPT, GPT-3.5, and GPT-4, in detecting sarcasm in natural language. It tests fine-tuned and zero-shot models of different sizes and releases.

The GPT models were tested on the political and balanced (pol-bal) portion of the popular Self-Annotated Reddit Corpus (SARC 2.0) sarcasm dataset. In the fine-tuning case, the largest fine-tuned GPT-3 model achieves accuracy and $F_1$-score of 0.81, outperforming prior models. In the zero-shot case, one of GPT-4 models yields an accuracy of 0.70 and $F_1$-score of 0.75. Other models score lower. Additionally, a model's performance may improve or deteriorate with each release, highlighting the need to reassess performance after each release.
 
\end{abstract}

\section{Introduction}\label{sec:intro}
Both humans and artificial intelligence have difficulty interpreting sarcasm correctly. It is especially challenging for textual inputs where body language and speaker intonation are absent. Sarcasm-detecting agents (i.e., systems that detect sarcasm in texts) are tested on their ability to interpret context when determining whether a textual statement is sarcastic~\citep{kruger2005egocentrism}.

The development of an accurate sarcasm detection systems holds great potential for improving human-computer interactions, given that sarcasm is widely used in human conversations~\citep{olkoniemi2016individual}. To detect sarcasm in text-based social interactions, a model with contextual knowledge and social understanding capabilities is needed. There has been extensive work in detecting sarcasm, especially with the Self-Annotated Reddit Corpus (SARC 2.0) dataset~\citep{khodak2017large}. Sarcasm detection models with the highest performance rely on Transformers~\citep{potamias2020transformer, fuzzy, commonsense}, recurrent neural networks~\citep{bilstm, potamias2020transformer, ilic-etal-2018-deep}, and/or feature engineering~\citep{cascade, commonsense}.

OpenAI Generative Pre-trained Transformer (GPT) models have shown effectiveness in natural language understanding tasks~\citep{chen2023chatgpt}. However, to our knowledge, there is no comparative study of fine-tuning and zero-shot methods for sarcasm detection, and no study of differently versioned ChatGPT models' sarcasm detection abilities.

This paper aims to fill this gap by analyzing the performance of GPT models in detecting sarcasm using the SARC 2.0 political, balanced (pol-bal) dataset. 
For the sake of brevity, from hereon, we will refer to this dataset as \textit{pol-bal} dataset.
Our research questions (RQs) are as follows.
\begin{enumerate}[label=\textbf{RQ\arabic*.}]
    \item How does model size affect the ability of fine-tuned GPT-3 models to detect sarcasm?
    \item What are the characteristics of the top-performing zero-shot GPT model?
    \item How is zero-shot learning affected by different versions of the same GPT model? 
    \item How is fine-tuned learning affected by different versions of the same GPT model? 
\end{enumerate}
Our key \textbf{contribution} is analyzing how different model sizes, versions, and learning methods influence GPT models' ability to detect sarcasm.

The rest of this article is structured as follows. Section~\ref{sec:lit_review} presents a literature review. Section~\ref{sec:methodology} covers the methodology of our experiments, while Section~\ref{sec:results} discusses the results. Finally, Section~\ref{sec:summary} concludes the paper.

\section{Literature Review}\label{sec:lit_review}
The following literature review aims to provide an overview of popular sarcasm detection datasets, research on the \textit{pol-bal} dataset, implications of Large Language Models (LLMs) on sarcasm detection, and GPT models’ abilities for other natural language understanding (NLU) tasks.

\subsection{Sarcsam Detection Datasets}\label{sec:datasets}
Multimodal Sarcam Detection Dataset (MUStARD), developed by \cite{castro-etal-2019-towards}, consists of 690 (50\% sarcastic) scenes from four television shows with situational context given in the form of respective speakers and previous dialogue. It intends to challenge multi-modal models with sarcasm detection given situational context. However, the dataset may also be used for text-based classification. 

The iSarcasmEval dataset \citep{abu-farha-etal-2022-semeval} contains 6,135 (21\% sarcastic) English tweets from English speakers asked to provide their own sarcastic and non-sarcastic tweets. It also provides self-annotated Arabic sarcastic statements. This SemEval task consists of three subtasks: sarcasm detection, sarcasm categorization, and pairwise sarcasm identification. 

Providing 362 (50\% sarcastic) statements, SNARKS \citep{snarks} uses a contrastive minimal-edit distance (MiCE) setup, presenting a binary choice sarcasm detection task. SNARKS omits sarcastic statements requiring factoid-level knowledge as well as the comment thread leading to a sarcastic statement. SNARKS's sarcasm detection challenge with a MiCE setup tests a model's understanding of a word's literal meaning and its knowledge of its features without considering its context.

\subsubsection{Dataset under study: SARC 2.0 \textit{pol-bal}}\label{sec:sarc_pol_bal}

SARC is a large dataset containing 1.3 million sarcastic Reddit comments from various sub-reddits. A popular benchmark is a subset of this dataset that contains a balanced sample of sarcastic and non-sarcastic comments from the r/politics sub-reddit. This subset is denoted \textit{pol-bal}; it poses the challenge for a model to understand sarcasm within a real social interaction|with situational context|while considering background factoid-based political knowledge. Evaluation on \textit{pol-bal} dataset offers insight into a model's ability to understand sarcasm in a real-world scenario. When detecting sarcasm, especially when a particular political background is required, the researcher's model must be robust.

An interesting feature of the \textit{pol-bal} dataset is its balancing method, where each observation includes a comment thread and two replies to said thread: one sarcastic and one non-sarcastic. Thus, both train and test data are ideally balanced, with 50\% of observations being sarcastic and 50\% not sarcastic.  The \textit{pol-bal} dataset contains 13,668 training and 3,406 testing observations. Each sarcastic comment in the dataset is given one to six preceding comments in its respective thread from the forum.

We have chosen to utilize \textit{pol-bal} dataset due to its unique challenges, high amount of observations, and popularity. 

\subsection{Research on \textit{pol-bal} dataset}\label{sec:research_sarc_pol_bal}
The following models have been constructed to detect sarcasm in SARC \textit{pol-bal} dataset\footnote{\citet{commonsense} propose a method for detecting sarcasm by addressing the notion that sarcasm detection is based on common sense. Their method is based on using BERT- and COMET-generated related knowledge. This method yields accuracy $\approx 0.76$ and $F_1 \approx 0.76$. They are using another political subset of SARC 2.0. Thus, we cannot compare their performance with that of other papers.

\cite{choi2023llms} create a benchmark for social understanding for LLMs; they group social understanding tasks into five categories, including humour and sarcasm. The SARC 2.0 dataset is included in theirs. A DeBERTa-V3 model is found to be the best at detecting sarcasm among the other LLMs tested. However, they do not use the \textit{pol-bal} subset, so the results are not comparable.

\citet{fuzzy} leverage BERT and fuzzy logic to detect sarcasm. Despite the fact that they present performance results for SARC 2.0, the data subset in their study mixed pol and main SARC 2.0 data subsets, making it impossible to compare their results to other papers in this section. 
}. 
Summary of the models' performance is given in Table~\ref{tbl:prior_models}. 

\citet{khodak2017large}  classify their \textit{pol-bal} test set using several models. The best result uses a Bag-of-Bigrams approach and achieves accuracy $\approx 0.77$.

It is not trivial to detect sarcasm, as stated above. According to \citep{khodak2017large}, five human ``labelers'' attain an average accuracy $\approx 0.83$. A majority vote among the ``labelers'' improves accuracy to $\approx 0.85$.

\citet{cascade} develop the ContextuAl SarCasm DEtector (CASCADE) model which uses both content and context modeling to classify an r/politics post’s reply as sarcastic.
They use a combination of convolutional neural and feature embedding techniques to model the data.
This method reaches an accuracy $\approx 0.74$ and $F_1 \approx 0.75$.

\citet{pelser2019deep} use a dense and deeply connected model in an attempt to extract low-level features from a sarcastic comment without the inclusion of its respective situational context. This method manages to attain accuracy $\approx 0.69$ and $F_1 \approx 0.69$.
 
\citet{ilic-etal-2018-deep} attempt to classify sarcasm based on morpho-syntactic features of a sarcastic statement. This method uses ELMo word embeddings passed through a BiLSTM to classify sarcasm. It achieves accuracy $\approx 0.79$.

\citet{potamias2020transformer} propose a method called RCNN RoBERTa, which uses RoBERTa embeddings fed into a recurrent convolutional neural network to detect sarcasm. This method achieves an accuracy $\approx 0.79$ and $ F_1 \approx 0.78$.

Thus, two methods achieve state-of-the-art accuracy \citep{ilic-etal-2018-deep, potamias2020transformer}.

\begin{table}[t]
\caption{Performance of models on SARC \textit{pol-bal} dataset. A -- indicates cases in which $F_1$ are not reported.}\label{tbl:prior_models}
\begin{center}
\begin{tabular}{@{}llrr@{}}
\toprule
Reference                      & Model                       & Acc & $F_1$   \\ \midrule
\cite{pelser2019deep}          & dweNet                      & 0.69  & 0.69 \\ 
\cite{cascade}                 & CASCADE                     & 0.74  & 0.75 \\ 
\cite{khodak2017large}         & Bag-of-Bigrams              & 0.77  &   --   \\ 
\cite{potamias2020transformer} & RCNN-RoBERTa                & 0.79  & 0.78 \\ 
\cite{ilic-etal-2018-deep}     & ELMo-BiLSTM                 & 0.79  &  --    \\ 
\cite{khodak2017large}         & Human (Average)          & 0.83  &   --   \\ 
\cite{khodak2017large}         & Human (Majority)          & 0.85  &  --    \\ \bottomrule
\end{tabular}
\end{center}
\end{table}

\subsection{Research on sarcasm detection with LLMs}
\cite{srivastava2023imitation} create the BIG-bench benchmark for natural language understanding tasks with large language models. It details the usage of eight differently-sized OpenAI GPT-3 models from \citep{brown2020language} on 204 natural language tasks in a zero-shot and few-shot manner. In particular, they conduct $[0, 1, 2, 5]$-shot testing on GPT-3 with observations from the SNARKS sarcasm dataset (discussed in Section~\ref{sec:datasets}). As they are using a different dataset, their work is complementary to ours. 

\cite{mu2023navigating} compare the sarcasm-detecting abilities of GPT-3.5-turbo, OpenAssistant, and BERT-large on the iSarcasmEval dataset  (discussed in Section~\ref{sec:datasets}). They find that BERT-large is the most performative on this dataset. This work is complementary to ours as it uses a different dataset and a single GPT model.

\cite{choi2023llms} create a benchmark for social understanding for LLMs; they group social understanding tasks into five categories, including humor and sarcasm, which includes the SARC 2.0 dataset. As discussed in Section~\ref{sec:research_sarc_pol_bal}, their work is complementary to ours.

\cite{gptsarcasm} use a fine-tuned GPT-3 curie and zero-shot text-davinci-003 models on the MUStARD dataset (discussed in Section~\ref{sec:datasets}) and achieve top $F_1=0.77$. This work is complementary to ours as they are using a different dataset.

\subsection{Research on other Natural Language Understanding tasks using GPT models}
\cite{chen2023robust} compare $[0, 1, 3, 5]$-shot LaMDA-PT, $[0, 2, 3, 4, 10, 15, 16]$-shot FLAN, and fine-tuned popular approaches  to $[0, 3, 6, 9]$-shot InstructGPT models (text-davinci-001, text-davinci-002) and the text-davinci-003 GPT-3.5 model on 21 datasets across 9 different NLU tasks not including sarcasm detection. They find that the GPT-3.5 model performs better than the other models in certain tasks like machine reading comprehension and natural language inference, but performs worse than other models in sentiment analysis and relation extraction.

\section{Methodology}\label{sec:methodology}

Our dataset under study---\textit{pol-bal}---is discussed in Section~\ref{sec:sarc_pol_bal}. 
Using this dataset, we study fourteen versions of GPT models shown in Table~\ref{models_studies}. OpenAI GPT are generative models pre-trained on a large corpus of text data~\citep{brown2020language}. GPT models are pre-trained to predict the next token in a given document, learning to estimate the conditional probability distribution over its vocabulary given the context; see~\citet{zhao2023survey} for review. With pre-training, GPT models are equipped with a vast amount of language knowledge and world information, which, in conjunction with their large parameter count, allows them to excel at natural language tasks~\citep{brown2020language}. 

The four model families tested in this paper are GPT-3, InstructGPT, GPT-3.5, and GPT-4. Technical details of the models are given in~\cite{brown2020language, ouyang2022training, chatGPTpost, openai2023gpt4}. In short, InstructGPT models are fine-tuned GPT-3 models aligned with human intent \citep{ouyang2022training}. The GPT-3.5 models, like text-davinci-003, are based on code-based GPT-3 models \citep{zhao2023survey}. GPT-3.5-turbo is a version of GPT-3.5 trained to follow instructions and give detailed responses; it is a ChatGPT model~\citep{chatGPTpost}. GPT-4 is currently the most successful GPT model for professional and academic tasks; it is also a ChatGPT model~\citep{openai2023gpt4}.

\begin{table}[t]
\caption{GPT models under study. * denotes ChatGPT model. The size of the ada, babbage, curie, and davinci models is reported by~\citet{brown2020language}. The size of the remaining models is unknown. However, it is conjectured that the size of InstructGPT and GPT-3.5 models is similar in magnitude to their predecessors~\citep{zhao2023survey}, while the size of GPT-4 is larger than of their predecessors~\citep{schreiner2023gpt}. }
\label{models_studies}
\begin{center}
\begin{tabular}{@{}llrr@{}}
\toprule
Model Family & Version                & \# of Parameters & Release Date  \\ \midrule
GPT-3        & ada                    & 0.4 B      & June 2020        \\
GPT-3        & babbage                & 1.3 B        & June 2020             \\
GPT-3        & curie                  & 6.7 B       & June 2020                \\
GPT-3        & davinci                & 175.0 B      & June 2020            \\
InstructGPT & text-ada-001     & unknown          &  January 2022              \\
InstructGPT & text-babbage-001 & unknown          &    January 2022                       \\
InstructGPT & text-curie-001   & unknown        &          January 2022                \\
GPT-3.5      & text-davinci-003 & unknown          &     January 2022                    \\
GPT-3.5      & GPT-3.5-turbo-0301*         & unknown          & March 2023                  \\
GPT.3.5      & GPT-3.5-turbo-0613*         & unknown          & June 2023                   \\
GPT.3.5      & GPT-3.5-turbo-1106*         & unknown          & November 2023                   \\
GPT-4        & GPT-4-0314*             & unknown          & March 2023                \\
GPT-4        & GPT-4-0613*             & unknown          & June 2023                    \\ 
GPT-4        & GPT-4-1106-preview*             & unknown          & November 2023                    \\ 
\bottomrule
\end{tabular}
\end{center}

\end{table}

To answer this work's research questions, we initially developed prompts which wrap each observation from the dataset. Subsequently, we fine-tuned and tested all GPT-3 and some GPT-3.5 models; zero-shot tested the GPT-3, InstructGPT, GPT-3.5, and GPT-4 models; and finally performed statistical analyses on their results as discussed below.

\subsection{Prompt Development Stage}
This sarcasm classification problem was addressed by creating two prompts: one to wrap input data and a system prompt for zero-shot testing (shown in Figure~\ref{fig:prompts}). 

We set \texttt{<completion-delimiter>} placeholder to \verb|\\n\\n###\\n\\n| when fine-tuning GPT-3 models. And when performing the rest of the experiments, we set the placeholder to \verb|\n\n###\n\n|.

During fine-tuning and at inference, the \texttt{<thread>} placeholder was replaced with a given observation's context (the thread of text leading up to the response), while the \texttt{<reply>} placeholder was replaced with an observation's response. 

We introduced the \texttt{\textbackslash{n}Post\_n:} delimiter between comments in cases where multiple comments were provided within a given observation (as discussed in Section~\ref{sec:sarc_pol_bal}). An example of such an observation is given in Figure~\ref{fig:multi_post_example}.

\begin{figure}[tb]
     \begin{subfigure}[b]{0.4\textwidth}
         \centering
         \fbox{\begin{minipage}{\linewidth}
        \begin{alltt}
        Thread: <thread>\textbackslash{n}\textbackslash{n}Reply: <reply><completion-delimiter>
        
        \end{alltt}
        \end{minipage}}

         \caption{Wrapping prompt.}
         \label{fig:prompt_wo_env}
     \end{subfigure}
     \hfill
     \begin{subfigure}[b]{0.5\textwidth}
         \centering
         \fbox{\begin{minipage}{\linewidth}
            \begin{alltt}
            Classify each comment thread's response as sarcastic with yes or no.\textbackslash{n}
            \end{alltt}
            \end{minipage}}
                     \caption{Zero-shot prompt's prefix.}
         \label{fig:prompt_zero_shot}
     \end{subfigure}
        \caption{Prompts used for fine-tuning and zero-shot GPT models. \texttt{<reply>} and \texttt{<thread>} placeholders represent template values that need to be replaced with actual content.}
        \label{fig:prompts}
\end{figure}

\begin{figure}[tb]
     \centering

         \fbox{\begin{minipage}{\linewidth}
            \begin{alltt}
            Classify each comment thread's response as sarcastic with yes or no.\textbackslash{n}Thread: 'Post\_0: '\textbf{\textit{Just finished watching the debate. I love the President!}}'\textbackslash{n}Post\_1: '\textit{\textbf{Agreed! Can't wait for the next event!}}''\textbackslash{n}\textbackslash{n}Reply: '\textit{\textbf{Oh, the suspense is killing me!}}'\textbackslash{n}\textbackslash{n}\#\#\#\textbackslash{n}\textbackslash{n}
            \end{alltt}
            \end{minipage}}
        \caption{A zero-shot test prompt with two comments in a thread.}
        \label{fig:multi_post_example}
\end{figure}

\subsection{Fine-tuning stage}\label{sec:method_fine_tune}

\paragraph{GPT-3}
To test how the model size affects a fine-tuned GPT model’s ability to detect sarcasm, fine-tuned versions of all the GPT-3 models were created and tested. 

\paragraph{GPT-3.5}
At the time of writing OpenAI does not allow fine-tuning of the GPT-3.5 text-davinci-003 and gpt-3.5-turbo-0314 models~\citep{gpt35_fine_tuning}. 

GPT-3.5-turbo-0613 and GPT-3.5-turbo-1106 models were fine-tuned, but OpenAI does not report their size. Thus, we cannot use them to answer RQ1, but we can use them to answer RQ4.

\paragraph{GPT-4}
When the experiments were conducted, OpenAI did not allow fine-tuning GPT-4 models. OpenAI will enable fine-tuning of GPT-4 models later in the year~\citep{gpt35_fine_tuning}. We plan to assess the performance of GPT-4 fine-tuned models in the future. 

\paragraph{Hyperparameters}
The fine-tuning was performed using the following hyperparameter values: a learning rate of 0.1, 4 epochs, a prompt loss weight of 0.01 (applicable only to GPT-3 models), and a batch size of 16. Each model was trained using the prompt shown in Figure~\ref{fig:prompt_wo_env}. 

\paragraph{Training and validation datasets}
Each fine-tuned model was trained on the \textit{pol-bal} train set, with its desired completion being ``~yes" or ``~no". The training-validation split for each fine-tuned model had a ratio of 75\% for training and 25\% for validations (shuffled at random once and then reused for each fine-tuning session). Once fine-tuned, a model is tested on the \textit{pol-bal} test set with its temperature set to zero (in order to reduce the variability of the model's output) and restricted to outputting one token. 

\subsection{Zero-shot Stage}\label{sec:method_zero_shot}
Zero-shot testing methods were used to explore further how different generations and re-releases, impact a GPT model's sarcasm detection performance.

The prompt in Figure~\ref{fig:prompt_wo_env} was prefixed with the sentence in Figure~\ref{fig:prompt_zero_shot} during inference on zero-shot models|and added as a system prompt for ChatGPT models (this system prompt is identical to the zero-shot prefix without a tailing \texttt{\textbackslash{n}}). Thus, the final prompt asks the model to answer ``yes" or ``no". In practice, the models return complete sentences containing ``yes" and ``no" words\footnote{The number of output tokens is set to 128.}. Model output is mapped to a sarcastic or not sarcastic label using a regular expression that checks for ``yes" or ``no".

Some models may also return empty outputs or corrupted sentences without ``yes'' or ``no'' keywords (or their synonyms). The logit bias was introduced to the models to force them to return only ``yes'' or ``no'' tokens. 

The output of the models with and without logit bias is compared. Also reported are the number of observations that did not return ``yes'' or ``no'' keywords.

\subsection{Data analysis stage}\label{sec:method_data_analysis}

Accuracy and $F_1$-score metrics are used to measure the performance of classification methods. The metrics we used are the same as those used by other researchers studying this dataset (see Section~\ref{sec:sarc_pol_bal} for details). From the metrics perspective, the sarcastic label is considered \textit{positive}, whereas the non-sarcastic label is considered \textit{negative}.

To establish a baseline, we can use the naive ZeroR classifier, which labels every test observation as sarcastic. Given that the dataset is balanced, ZeroR classifier accuracy $= 0.5$ and $F_1 \approx 0.67$.

As described in Section~\ref{sec:method_zero_shot}, some models cannot label some observations. Such observations are removed from the list of observations used to calculate accuracy and $F_1$-score.

The exact version of McNemar's test~\citep{mcnemar1947note} is used to determine if there was a statistically significant difference between the classification methods' performance, as recommended by~\cite{dietterich1998approximate}.

\section{Results}\label{sec:results}

Below are the results of the experiments described in Section~\ref{sec:methodology}. Sections~\ref{sec:rq_size}--\ref{sec:fine_tuned_version} present the results as answers to our four research questions (defined in Section~\ref{sec:intro}). Section~\ref{sec:discussion} discusses the results and identifies and identify possible threats to their validity.

\subsection{RQ1: How does model size affect the ability of fine-tuned GPT-3 models to detect sarcasm?}\label{sec:rq_size}
    
The \textbf{answer to RQ1} is as follows. In Table~\ref{tbl:results-fine-tuned}, the accuracy and $F_1$-scores of fine-tuned models increased monotonically with model size. Using the smallest model~--- ada~--- produces the worst results, while using the largest model~--- davinci~--- produces the best results. To the best of our knowledge, a fine-tuned davinci model achieves state-of-the-art results (accuracy $\approx 0.81$ and $F_1 \approx 0.81$) when applied to the \textit{pol-bal} dataset. 
To ensure that the difference in performance was statistically significant, we performed pairwise analysis using McNemar's test for all fine-tuned models. According to the test, the differences between all the models are statistically significant ($p < 0.05$), see Appendix~\ref{sec:mcnemar_fine_tuned} for details.

\begin{table}[tb]
\caption{Results of classification of the \textit{pol-bal} for fine-tuned models. We highlight the best results for fine-tuned models in \textbf{bold}. A column labeled Missed shows the percentage of observations (out of 3406 test observations) where a model returned no labels. }\label{tbl:results-fine-tuned}
\centering
\begin{tabular}{@{}llrrrr@{}}
\toprule
Model Family       & Model              & Parameters    & \multicolumn{3}{c}{Performance} \\ \cmidrule(lr){4-6}
   &                                    & count         & Acc           & $F_1$         & Missed (\%) \\ \midrule
\multirow{4}{*}{GPT-3}   & ada                & 0.4 B   & 0.738 & 0.737 & 0.00\% \\
                         & babbage            & 1.3 B   & 0.755 & 0.755 & 0.00\% \\
                         & curie              & 6.7 B    & 0.781 & 0.784 & 0.00\% \\
                         & davinci            & 175.0 B & \textbf{0.810} & \textbf{0.808} & 0.00\% \\ \midrule
\multirow{2}{*}{GPT-3.5} & gpt-3.5-turbo-0613 & unknown & 0.776 & 0.786 & 0.00\% \\
                         & gpt-3.5-turbo-1106 & unknown & 0.781 & 0.803 & 0.00\% \\ \bottomrule
\end{tabular}
\end{table}

\subsection{RQ2: What are the characteristics of the top-performing zero-shot GPT model?}

Table~\ref{tbl:results-zero-shot} shows that the GPT-3, InstructGPT, and GPT-3.5 text-davinci-003 models perform worse than (or comparable to) the ZeroR classifier (described in Section~\ref{sec:method_data_analysis}). Although logit bias reduces the count of missing observations, low accuracy and $F_1$-scores indicate that the models cannot differentiate between sarcastic and non-sarcastic comments. Even for top-performing models (e.g., GPT-4), the addition of logit bias leads to performance degradation.

Some versions of GPT-3.5-turbo and GPT-4 models perform better than the ZeroR baseline. However, except for GPT-4 gpt-4-0613 and gpt-4-1106-preview, their performance is lower than those of the simpler models shown in Table~\ref{tbl:prior_models}.

The zero-shot models ``challenge'' is won by GPT-4 GPT-4-0613 model, achieving accuracy~$\approx 0.70$ and $F_1 \approx 0.75$. This model had no missing observations.
Note that gpt-4-1106-preview achieves higher accuracy~$\approx 0.72$ and lower $F_1 \approx 0.74$, but this model answered ``yes" or ``no" only to $99.97$\% of observations.

Based on this analysis, the \textbf{answer to RQ2} is as follows. In the \textit{pol-bal} dataset, only the most sophisticated GPT model (i.e., GPT-4 GPT-4-0613) can detect sarcasm competitively using the zero-shot approach. However, Table~\ref{tbl:prior_models} shows that GPT-4-0613 model performs poorly in comparison with prior simpler models, placing it second-last among them.

\begin{table}[tb]
\centering
\caption{Results of classification of the \textit{pol-bal} for zero-shot models. We highlight the best results for fine-tuned models in \textbf{bold}. A column labeled Missed shows the percentage of observations (out of 3406 test observations) where a model returned no labels.}
\label{tbl:results-zero-shot}
\begin{tabular}{ll|rrr|rrr}
\toprule
Model Family       & Model              & \multicolumn{3}{c|}{Performance (w/o bias)} & \multicolumn{3}{c}{Performance  (w bias)} \\ \cmidrule(lr){3-5} \cmidrule(lr){6-8}
   &                                    & Acc   & $F_1$ & Missed (\%) &  Acc   & $F_1$ & Missed (\%) \\ \midrule
\multirow{4}{*}{GPT-3}       & ada                & 0.400 & 0.000 & 99.85 & 0.500 & 0.666 & 0.00 \\
                             & babbage            & 0.333 & 0.000 & 99.82 & 0.500 & 0.000 & 0.00 \\
                             & curie              & 0.607 & 0.645 & 99.18 & 0.500 & 0.000 & 0.00 \\
                             & davinci            & 0.526 & 0.076 & 95.48 & 0.509 & 0.658 & 0.00 \\ \midrule
\multirow{3}{*}{InstructGPT} & text-ada-001       & 0.508 & 0.646 & 37.52 & 0.500 & 0.604 & 0.00 \\
                             & text-babbage-001   & 0.463 & 0.541 & 32.47 & 0.484 & 0.503 & 0.00 \\
                             & text-curie-001     & 0.501 & 0.667 & 0.15  & 0.500 & 0.667 & 0.00 \\ \midrule
\multirow{4}{*}{GPT-3.5}     & text-davinci-003   & 0.467 & 0.266 & 0.00  & 0.479 & 0.404 & 0.00 \\
                             & gpt-3.5-turbo-0301 & 0.592 & 0.653 & 0.03  & 0.587 & 0.656 & 0.00 \\
                             & gpt-3.5-turbo-0613 & 0.500 & 0.002 & 0.00  & 0.499 & 0.032 & 0.00 \\
                             & gpt-3.5-turbo-1106 & 0.578 & 0.613 & 0.00  & 0.504 & 0.581 & 0.00 \\ \midrule
\multirow{3}{*}{GPT-4}       & gpt-4-0314         & 0.599 & 0.710 & 0.00  & 0.601 & 0.712 & 0.00 \\
                             & gpt-4-0613         & \textbf{0.701} & \textbf{0.748} & \textbf{0.00}  & 0.680 & 0.738 & 0.00 \\
                             & gpt-4-1106-preview & 0.717 & 0.739 & 0.03  & 0.694 & 0.744 & 0.00 \\ \bottomrule
\end{tabular}
\end{table}

\subsection{RQ3: How is zero-shot learning affected by different versions of the same GPT model? }\label{sec:rq4}

In our work, we have two models with three versions, namely 
\begin{enumerate}
    \item GPT-3.5-turbo released in March (0301), June (0614), and November (1106) of 2023 and
    \item GPT-4 released in March (0314), June (0614), and November (1106-preview) of 2023.
\end{enumerate}

McNemar's test $p$-values for these models are provided in Appendix~\ref{sec:mcnemar_versions}. According to the results, the difference between most releases is statistically significant ($p < 0.05$). 

However, there are some exceptions. In the absence of bias, according to McNemar's test, gpt-3.5-turbo-0301 yields results similar to gpt-3.5-turbo-1106 and gpt-4-0314. 
With bias, gpt-3.5-turbo-0301 is similar to gpt-4-0314, and gpt-3.5-turbo-0613 is similar to gpt-3.5-turbo-1106.
However, the differences in accuracy and $F_1$ values produced by these models may suggest that this similarity is an artifact of statistics.

As shown in Table~\ref{tbl:results-zero-shot}, with bias, accuracy and $F_1$ increased monotonically for GPT-4. Every new GPT-3.5-turbo model has not reached the performance of its first release (gpt-3.5-turbo-0301). A bias-free comparison is more challenging because some models lack observations, but it seems that the latest models are not the best.

In other words, \textbf{the answer to RQ3} is as follows. The GPT model's ability to detect sarcasm may decline or improve with new releases, as was observed for other tasks by~\cite{chen2023chatgpt}.

\subsection{RQ4: How is fine-tuned learning affected by different versions of the same GPT model? }\label{sec:fine_tuned_version}
According to Table~\ref{tbl:results-fine-tuned}, both GPT-3.5 models (gpt-3.5-turbo-0613 and gpt-3.5-turbo-1106) perform similarly, although a newer model exhibits slightly better performance, with accuracy rising from 0.776 to 0.781 and $F_1$-score increasing from 0.786 to 0.803. According to McNemar's test (shown in Appendix~\ref{sec:mcnemar_fine_tuned}), this difference is not statistically significant.

Fine-tuned GPT-3.5 models perform worse than GPT-3 davinci models. McNemar's test reveals that the performance of both GPT-3.5 models is comparable (from a statistical perspective) to that of GPT-3 curie. Since OpenAI does not report the size and architecture of GPT-3.5 models, it is difficult to draw strong conclusions from this observation.

\textbf{The answer to RQ4} is as follows. Versioning of fine-tuned models does not have a significant impact on the model's performance. Newer versions of the models may change the answer in the future.

\subsection{Discussion and threats to validity}\label{sec:discussion}
This article groups threats to validity into four categories (internal, construct, conclusion, and external) as presented in~\cite{yin2017case}.
 
\paragraph{Internal validity}
In order to reduce the risk of errors, all experiments and data analysis were automated using Python scripts (which were cross-reviewed and validated by the authors).
 
\paragraph{Construct validity}
This study uses accuracy and $F_1$-score to measure models' performance. Previous authors studying this dataset used the same metrics (see Section~\ref{sec:sarc_pol_bal} for details).
 
In this study, several formulations of prompts were tested; the most effective ones were reported. Our prompts may be suboptimal, and we encourage the community to create better ones. The performance of the models may be further enhanced with such better prompts.
 
Furthermore, our research scope did not cover sophisticated prompt engineering techniques, such as $k$-shot~\citep{brown2020language}, chain of thought with reasoning~\citep{kojima2022large}, or retrieval augmented generation~\citep{lewis2020retrieval} that could improve the models' performance. However, this study aims not to achieve the best results for sarcasm detection but to examine how various attributes of models and prompts affect classification performance.
 
Another threat is that we use a regular expression to map the output of models to a label without looking at the output context. Thus, during zero-shot tests, regular expressions may misclassify specific outputs. We have sampled and eyeballed many outputs to mitigate this risk. We observed that the keyword ``yes" mapped to ``sarcastic" and the keyword ``no" mapped to ``non-sarcastic" in all cases. Furthermore, if neither label ``yes" nor ``no" was present, the output was meaningless and could not be categorized.

As mentioned in Section~\ref{sec:method_fine_tune}, fine-tuning of GPT-4 models is not available at this time. We plan to fine-tune GPT-4 models when they become available in the future.

\paragraph{Conclusion validity}
 
We do not know the exact content of the datasets used to train GPT models~\citep{brown2020language}. Therefore, it is possible that GPT models saw \textit{pol-bal} dataset during training. The fact that fine-tuning improves the results significantly over zero-shot experiments may suggest otherwise. At the very least, it may suggest that the models have ``forgotten'' this information to a significant extent.

\paragraph{External validity}

Our study is based on a single dataset. Despite its popularity in sarcasm detection studies, our findings cannot be generalized to other datasets. However, the same empirical examination can also be applied to other datasets with well-designed and controlled experiments. Furthermore, this research serves as a case study that can help the community understand sarcasm detection trends and patterns and develop better LLMs to detect complex sentiments, such as sarcasm.
 
Although the GPT-3 davinci fine-tuned model produced best-in-class prediction results, it does not imply that it will be used for those purposes in practice due to economic concerns. Due to the large number of parameters in LLMs, inference is expensive. The expense may be justified in business cases where sarcasm detection is critical.

\section{Summary}\label{sec:summary}
Humans and machine learning models have difficulty detecting sarcasm. This study examines fourteen GPT models' ability to detect sarcasm in the popular \textit{pol-bal} dataset. The results are as follows.

The fine-tuned GPT-3 davinci model with 175 billion parameters outperforms prior models with an accuracy and $F_1$-score of 0.81. Currently, fine-tuned GPT-3.5 models perform worse than the GPT-3 davinci model; this may change in the future. The performance of two fine-tuned GPT-3.5 models was similar and comparable to that of the GPT-3 curie model.

In the zero-shot case, only the two latest GPT-4 models could detect sarcasm with comparable performance to the prior models (achieving the second-last result). The performance of a model could improve or decline with a new release, which implies that every new release should be reevaluated.
 
Practitioners interested in incorporating sarcasm detection into their applications may find this work useful. It may also interest academics who wish to improve machine learning models' ability to detect complex language usage.

\bibliography{references}

\appendix
\section{Results of the pairwise McNemar's test for fine-tuned GPT-3 and GPT-3.5 models}\label{sec:mcnemar_fine_tuned}

\begin{figure}[tb]
\centering
\includegraphics[width=0.85\textwidth]{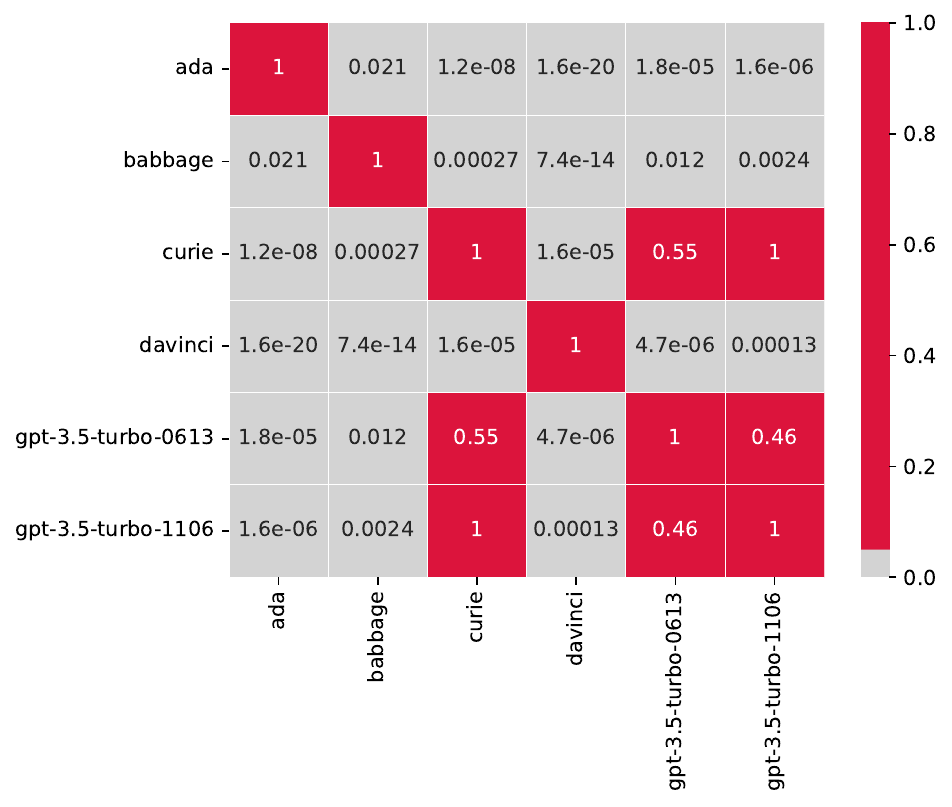}
\caption{P-values of the pairwise McNemar's test of fine-tuned GPT-3 and GPT-3.5-turbo models;  $p$-values $\ge 0.05$ are highlighted in red.}
\label{results-mcnemar-ft-noenv}
\end{figure}

McNemar's test~\citep{mcnemar1947note} is applied to predictions of each pair of classification models as per~\cite{dietterich1998approximate} to compare the performance of different-sized fine-tuned GPT-3 models. 
Figure~\ref{results-mcnemar-ft-noenv} show the results of the pairwise comparison of the models trained. 

Statistically significant differences are observed in all cases for GPT-3 models ($p < 0.05$). 

However, both GPT-3.5 models behave similarly ($p \ge 0.05$). Furthermore, their performance is comparable to that of GPT-3 curie!

\section{Results of the pairwise McNemar's test for different versions of GPT-3.5-turbo and GPT-4}\label{sec:mcnemar_versions}

\begin{figure}[tb]
\centering
\centering
\includegraphics[width=0.85\textwidth]{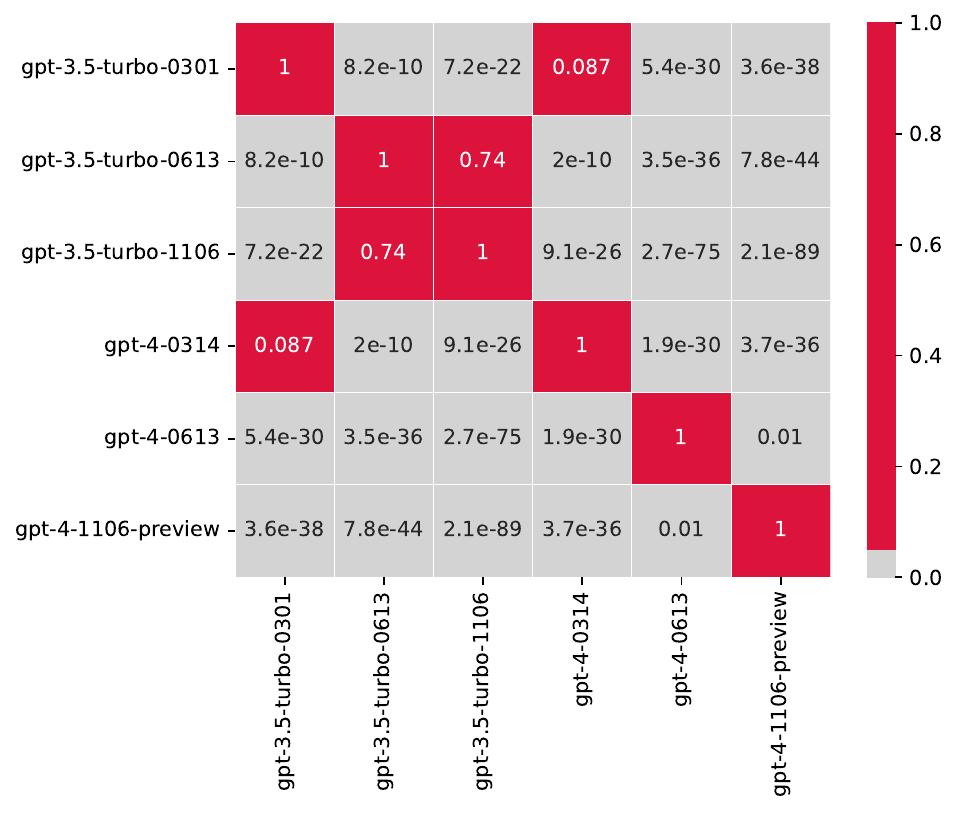}
\caption{P-values of the pairwise McNemar's test for zero-shot experiments with bias using  GPT-3.5-turbo and GPT-4 ChatGPT models; $p$-values $\ge 0.05$ are highlighted in red.}
\label{results-mcnemar-zeroshot-chat-nodomain-w-bias}
\end{figure}

\begin{figure}[tb]
\centering
\centering
\includegraphics[width=0.85\textwidth]{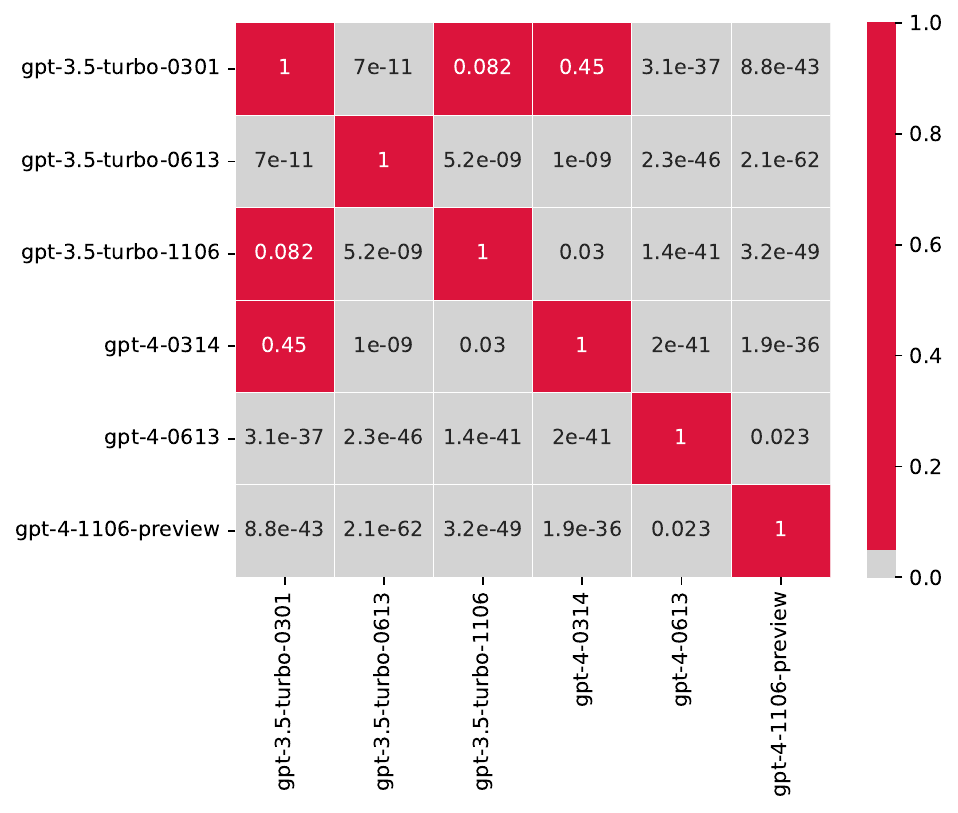}
\caption{P-values of the pairwise McNemar's test for zero-shot experiments without bias using GPT-3.5-turbo and GPT-4 ChatGPT models; $p$-values $\ge 0.05$ are highlighted in red.}
\label{results-mcnemar-zeroshot-chat-nodomain-wo-bias}
\end{figure}

In this section, we compare the difference in performance for different versions of GPT-3.5-turbo and GPT-4 models for the zero-shot case (with and without logit bias). McNemar's test~\citep{mcnemar1947note} is applied to predictions of each pair of classification models as per~\cite{dietterich1998approximate}. Figures~\ref{results-mcnemar-zeroshot-chat-nodomain-w-bias} and~\ref{results-mcnemar-zeroshot-chat-nodomain-wo-bias} show the pairwise comparison results of the models tested with and without bias, respectively.

\end{document}